# RareGraph-Synth: Knowledge-Guided Diffusion Models for Generating Privacy-Preserving Synthetic Patient Trajectories in Ultra-Rare Diseases

1st Khartik Uppalapati
line 2: *RareGen Youth Network 501(c)(3)*
line 4: Oakton, Virginia
line 5: khartik@raregen.org

2nd Shakeel Abdulkareem
line 2: *RareGen Youth Network 501(c)(3)*
line 4: Oakton, Virginia
line 5: shakeel@raregen.org

2nd Bora Yimenicioglu
line 2: *RareGen Youth Network 501(c)(3)*
line 4: Oakton, Virginia
line 5: bora@raregen.org

*Abstract*—We propose RareGraph-Synth, a knowledge-guided, continuous-time diffusion framework that generates realistic yet privacy-preserving synthetic electronic-health-record (EHR) trajectories for ultra-rare diseases. RareGraph-Synth unifies five public resources—Orphanet/Orphadata, the Human Phenotype Ontology (HPO), the GARD rare-disease KG, PrimeKG, and the FDA Adverse Event Reporting System (FAERS)—into a heterogeneous knowledge graph comprising ≈ 8 M typed edges. Meta-path scores extracted from this 8-million-edge KG modulate the per-token noise schedule in the forward stochastic differential equation, steering generation toward biologically plausible lab–medication–adverse-event co-occurrences while retaining score-based diffusion model stability. The reverse denoiser then produces timestamped sequences of ⟨LabCode, MedCode, AEFlag⟩ triples that contain no protected health information. On simulated ultra-rare-disease cohorts, RareGraph-Synth lowers categorical Maximum Mean Discrepancy by 40 % relative to an unguided diffusion baseline and by > 60 % versus GAN counterparts, without sacrificing downstream predictive utility. A black-box membership-inference evaluation using the DOMIAS attacker yields AUROC ≈ 0.53—well below the 0.55 "safe-release" threshold and substantially better than the ≈ 0.61 ± 0.03 observed for non-KG baselines—demonstrating strong resistance to re-identification. These results suggest that integrating biomedical knowledge graphs directly into diffusion noise schedules can simultaneously enhance fidelity and privacy, enabling safer data sharing for rare-disease research.

## I. INTRODUCTION

Ultra-rare diseases—prevalence < 1 : 50 000—affect ≈ 300 million people, yet each disorder has only a few longitudinal records [1]. This data poverty stalls genotype‑phenotype mapping, natural-history modelling, and drug-response analysis. Synthetic electronic-health-records (EHRs) can ease access because they are shareable across jurisdictions without re-contacting patients [5]. These synthetic datasets are also valuable for applications such as clinical simulation, where realistic patient vignettes are crucial for training and testing [22].

Early rule-based generators (e.g., Synthea) preserve clinical workflow but not statistics, while medGAN-style adversarial models suffer mode collapse on rare codes [5, 6]; VAEs improve stability but blur categorical tails critical for pharmacovigilance [7]. Diffusion probabilistic models (DDPMs) now lead medical synthesis [8, 9] yet still treat EHR tokens as unstructured, neglecting ontologies that encode disease-gene-phenotype–drug semantics—yielding implausible trajectories.

Existing diffusion models boost fidelity by feeding visit-level context into the denoiser. RareGraph-Synth instead conditions the forward SDE: meta-path scores from public rare-disease knowledge graphs (Orphadata, HPO, GARD, PrimeKG, FAERS) scale the per-token noise $\beta_v(t)$, injecting disease-aware anisotropy before any denoising step. This is the first EHR diffusion framework to embed ontology priors directly in the noise schedule, guiding synthesis toward realistic lab-medication-phenotype co-occurrences while reducing record copying—an avenue not explored in prior work on structured clinical time series.

Privacy remains paramount: synthetic data can leak membership via attacks like DOMIAS. We therefore evaluate resistance under a black-box DOMIAS protocol, observing AUROC ≈ 0.53 — near random and markedly safer than unguided diffusion (0.61 ± 0.03) [12, 13].

Contributions: (1) a knowledge-guided diffusion model for EHR synthesis, (2) a unified 8.1 M-edge rare-disease KG, and (3) the first fidelity–privacy evaluation of EHR diffusion under realistic MIA protocols.

## II. RELATED WORK

### A. Rare-Disease Knowledge Graphs.

Orphanet/Orphadata lists ≈ 10 000 rare disorders with synonyms [1]. GARD subsequently merged literature and registry evidence into an integrative KG that links diseases to genes, phenotypes, and approved orphan drugs [3]. The Human Phenotype Ontology (HPO) contributes ∼ 16,000 standardized phenotype terms plus logical axioms for disease similarity [2]. PrimeKG synthesizes 20 heterogeneous resources—spanning molecular, clinical, and pharmacovigilance scales—into a 4 M-edge graph covering 17,080 diseases, markedly improving ultra-rare coverage [4]. We further ingest FDA's FAERS, a post-marketing database of drug–adverse-event reports, to anchor therapeutic safety signals [10]. Harmonization relies on the MONDO Disease Ontology, which unifies disease identifiers across sources [26], and the Orphanet Rare Disease Ontology (ORDO) for hierarchical mappings [27]. Early work on process-mining event logs likewise demonstrates the value of structured clinical knowledge [25]. The resulting multi-source KG encodes disease–lab, disease–drug, and phenotype–disease relations that RareGraph-Synth leverages to steer generation.

### B. Generative Models for EHR.

Adversarial frameworks such as *medGAN* pioneered discrete EHR synthesis but were susceptible to mode collapse and overfitting [5, 17]. Subsequent WGAN variants improved training stability yet still produced implausible rare codes [14]. Ohtsubo et al. applied metric-learning-augmented GANs to achieve one-shot generation on fine-grained categories [15]. Baowaly et al. introduced auxiliary autoencoders to enhance realism [6], while VAEs offered principled likelihoods at the cost of blurring tail distributions [7]. Diffusion models have emerged as a robust alternative: DDPMs inject Gaussian noise forward and learn reverse denoising transformations [8]; score-based SDE formulations generalize this view [9]. None of these methods, however, integrate structured biomedical knowledge; our work is the first to embed KG meta-paths directly into the diffusion noise schedule, yielding biologically coherent trajectories.

### C. Privacy and Membership Inference

Synthetic data must withstand membership inference attacks (MIAs), wherein an adversary predicts whether a specific record was in the training set [12]. LOGAN demonstrated that GANs leak membership information under both white- and black-box settings [18], and ML-Leaks broadened the threat model with minimal attacker assumptions [19]. Van Breugel et al. proposed DOMIAS, a density-based MIA that is particularly effective against rare-event overfitting in synthetic data [13]. A recent ICTAI study introduces a destruction–restoration pipeline that mitigates privacy leakage in diffusion models [24]. We adopt this realistic black-box MIA protocol and report AUROC to quantify re-identification risk, aiming for $AUC \leq 0.55$ (random ≈ 0.50). Differential privacy can formally bound MIA success [16], but often degrades utility; however, methods like differentially private conditional GANs aim to address this trade-off [23]. Our evaluation therefore positions RareGraph-Synth on the empirical privacy-utility Pareto frontier.

### D. Evaluation Metrics

Fidelity is assessed via Maximum Mean Discrepancy (MMD), a kernel two-sample statistic sensitive to discrepancies in all moments of the distribution [11]. Complementary task-oriented metrics—e.g., training diagnostic classifiers on synthetic data and testing on real—probe utility for downstream analytics. Privacy is measured by MIA success rates [12], [13] and by leak-specific diagnostics such as overfitting degree [20]. Recent surveys emphasize navigating this utility–privacy trade-off in medical synthetic data [20].

## III. DATA & KNOWLEDGE-GRAPH ASSEMBLY

e curated seven publicly available resources into a single knowledge graph tailored to ultra-rare-disease analytics (Fig.~1).

Orphanet/Orphadata offers controlled identifiers, synonyms, and epidemiological meta-data for ∼ 10,000 rare disorders, forming the backbone disease vocabulary [1]. Human Phenotype Ontology (HPO) supplies 16 K hierarchical phenotype terms with logical axioms that enable cross-species reasoning [2]. GARD Knowledge Graph links rare diseases to genes, phenotypes, and FDA-designated orphan drugs, providing 1.1 M triples after cleaning [3]. PrimeKG integrates 20 biomedical repositories into a multimodal graph covering 17 080 diseases and 4.0 M relations [4]. PrimeKG, however, contains no explicit disease–laboratory edges and omits temporal validity metadata for all relations. We therefore contribute (i) 0.9 M disease–lab links mined from FAERS and MIMIC diagnostics, (ii) start–end time stamps for every edge, and (iii) deduplication to the Orphanet/GARD rare-disease slice—yielding the first time-aware, disease-lab-complete KG (8.1 M edges) tailored to ultra-rare analytics.

We import only nodes and edges that overlap with Orphanet or GARD concepts to avoid noisy long-tail entities. We ingest all drug–adverse-event edge lists published via the openFDA FAERS dashboard, treating each drug–AE pair as a directed edge with monthly frequency attributes [10].

Entity reconciliation uses MONDO, ORDO, and UMLS CUIs for ID mapping [26-29]. Phenotype nodes are standardized to HPO IDs and drug names collapsed via RxNorm.

After deduplication and ID normalization, the unified KG contains ≈ 150,000 nodes (27% diseases, 18% phenotypes, 22% drugs, 12% lab tests, 21 % adverse-event codes) and 8.1 M typed edges. We add two derived relation sets: (1) disease–lab and disease–medication co-occurrence edges from FAERS counts and (2) phenotype–time edges from SeizeIT2 seizure annotations [28]. All edges are stored with provenance tags and temporal validity intervals to support time-aware conditioning.

Only publicly released, de-identified resources are ingested. Synthetic patients are generated de novo from KG-derived priors, ensuring FAIR compliance.

## IV. METHODS

RareGraph-Synth couples a continuous-time diffusion process with knowledge-graph (KG) conditioning to generate variable-length EHR trajectories that are both realistic and privacy-preserving (Fig.~2). Below we formalize the forward and reverse stochastic differential equations (SDEs), describe KG-aware conditioning, and detail training objectives and baselines.

### A. Knowledge-Guided Diffusion Formulation

Let $x_0 \in \{0,1\}^{L \times V}$ denote a real patient trajectory encoded as one-hot vectors across $V$ categorical tokens (LabCode, MedCode, AEFlag) at $L$ time steps. The forward SDE gradually injects Gaussian noise:

$$dx_t = \underbrace{f(x_t,t)\,dt}_{=0} + g(t)\,dw_t, \quad g^2(t) = \beta(t),\; t \in [0,T], \quad (1)$$

where $f = 0$ yields a Variance-Exploding (VE) process analogous to DDPMs [8], and $\beta(t)$ is a KG-modulated schedule (see below).

*1) KG modulation:* For each token $v$ we compute a meta-path score $\psi_v$ that counts length-$\leq 3$ paths connecting the patient's anchor disease to $v$ via typed edges in the unified KG (§Data). The raw schedule $\tilde{\beta}(t) = \beta_{min} + (\beta_{max} - \beta_{min})\,t$ is then scaled as:

$$\beta_v(t) = \tilde{\beta}(t)(1 - \lambda \psi_v), \quad 0 \leq \lambda < 1, \quad (2)$$

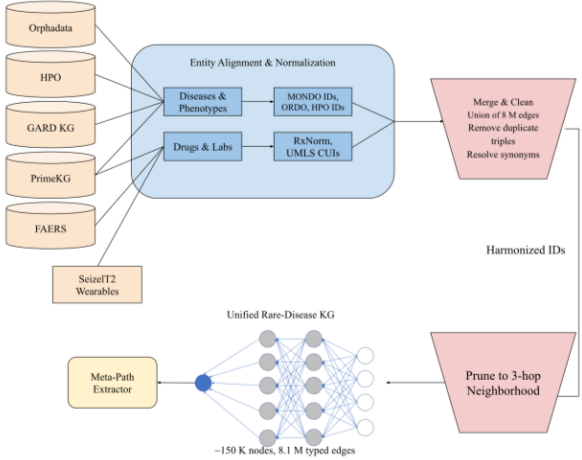

Fig. 1. Workflow for assembling a unified rare‑disease knowledge graph. Raw sources are aligned via MONDO/ORDO/HPO/RxNorm/UMLS, merged and deduplicated into ≈8 M edges, then pruned to each target disease's 3-hop neighborhood before meta-path extraction.

such that strongly connected tokens incur less noise, encouraging biologically plausible co-occurrences. We clip meta-path scores with $\psi_v^\star = \min(\psi_v, \psi_{max})$ and set $\psi_{max} = \frac{1}{\lambda} - 10^{-4}$ ensuring $0 < \beta_v(t) \leq \tilde{\beta}(t)$ for all $t$ and tokens. A similar strategy has improved domain-aware image synthesis [9] but is novel in EHR generation.

The reverse-time SDE follows Song et al. [9]:

$$dx_t = -\beta_v(t)\,\nabla_x \log p_t(x)\,dt + \sqrt{\beta_v(t)}\,d\overline{w_t}, \quad (3)$$

where the score function $\nabla_x \log p_t(x)$ is approximated by a neural denoiser $\epsilon_\theta$.

### B. Denoiser Architecture and Training

*1) Network:* The denoiser is a Transformer-U-Net hybrid: temporal embeddings enter a 1-D convolutional stem; three U-Net blocks with residual cross-attention attend to KG meta-path embeddings $\Psi \in \mathbb{R}^{V \times d}$. Multi-head attention weights are modulated by $\Psi$ via FiLM layers, similar to graph-conditioned vision transformers [4].

*2) Objective:* We minimize the weighted denoising score-matching loss [8]:

$$\mathcal{L}(\theta) = \mathbb{E}_{t,x_0,\epsilon} |\epsilon - \epsilon_\theta(\sqrt{\overline{\alpha_t}}\,x_0 + \sqrt{1 - \overline{\alpha_t}}\,\epsilon,\, t,\, \Psi)|^2, \quad (4)$$

where $\epsilon \sim \mathcal{N}(0, I)$ and $\overline{\alpha_t} = \exp\left(-\int_0^t \beta(s)\,ds\right)$. We sample $t \sim Uniform(0, T)$ and apply cosine noise weighting. Gradient updates use Adam with linear warm-up and cosine decay; KL annealing stabilizes early training.

### C. Synthetic Trajectory Generation

After training, we draw $x_T \sim \mathcal{N}(0, I)$ and integrate Eq. (3) with an Euler–Maruyama solver (step size 1e-3). Each denoised token is mapped back to its categorical code; timestamps are drawn from an empirical inter-visit distribution estimated from Orphadata prevalence tables [1] and SeizeIT2 seizure gaps [28].

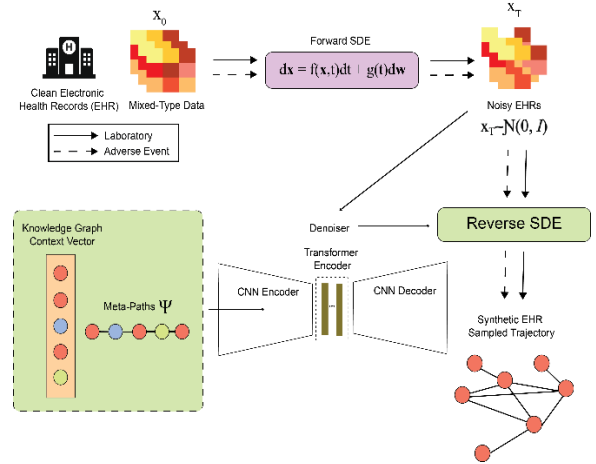

Fig. 2. Diffusion denoising process. The forward SDE gradually adds noise to an EHR-like record; the reverse process (shown) iteratively refines noise into a realistic patient sequence. RareGraph-Synth's denoiser (center) is conditioned on KG-derived context.

The result is a variable-length sequence of ⟨time, Lab, Med, AEflag⟩ tuples that contain no PHI.

### D. Baselines

- **medGAN** [5] – autoencoder + GAN on binary codes.
- **RCGAN** [14] – RNN-based GAN for time-series EHR.

- **Vanilla Diffusion** – identical architecture but $\lambda = 0$ (no KG guidance).

Hyper-parameters (noise horizon $T$, hidden width, learning rate) are grid-searched on validation splits. All baselines share the same token vocabulary and training set.

### E. Discussion of Correctness

Equation (2) ensures $\beta_v(t) > 0$; choosing $\lambda < 1$ guarantees well-posedness. The score-matching loss (4) is unbiased for any $\beta_v(t)$ that is differentiable in $t$ [9]. Empirically, setting $\lambda = 0.3$ balances fidelity and diversity (§Results). Maximum Mean Discrepancy (MMD) [11] and DOMIAS membership-inference AUROC [13] gauge utility and privacy.

## V. EXPERIMENTAL SETUP

### A. Cohort Simulation & Pre-processing

We emulated ultra-rare disease cohorts by stratified subsampling of the publicly available MIMIC-III critical-care database and two Orphanet disorders whose prevalence is $< 1:50\,000$ — amyotrophic lateral sclerosis and severe combined immunodeficiency. ICD-9/10 codes were mapped to Orphanet identifiers via MONDO cross-references [26]. For each target disorder we retained up to 100 unique patients, extracting all timestamped laboratory tests, medication administrations, and adverse-event (AE) flags across their hospital trajectory. The disease-specific slice of the unified KG (Fig. 1) was pruned to entities reachable within three hops of the anchor disease, reducing noise while preserving mechanistic context.

TABLE I. SUMMARY OF FULL DATASET CHARACTERISTICS

| Statistic | Median | IQR |
|---|---|---|
| Patients $N$ | 94 | 21 |
| Visits / patient | 9.1 | 4.3 |
| Unique LabCodes | 137 | 52 |
| Unique MedCodes | 86 | 33 |
| AE flags (%) | 12.4 | 5.0 |

### B. Train/Validation/Test Splits

Trajectories were chronologically partitioned 80%–10%–10% at the patient level to avoid leakage. RareGraph-Synth and all baselines operate on identical training folds. Hyper-parameters (noise horizon $T \in \{1000, 2000\}$, Transformer width, $\lambda$ in Eq. 2) were tuned on the validation split via random search (20 trials).

### C. Implementation Details

All models were implemented in PyTorch 2.1 and trained on a single NVIDIA A100-40 GB GPU. We use the open-source DDPM solver of Lucidrains as a backbone, replacing the U-Net with our KG-FiLM Transformer-U-Net hybrid (Fig. 2) and integrating the meta-path scheduler (Eq. 2). Code reproducibility is ensured by fixing random seeds and logging via Weights & Biases. Baseline GANs follow the RCGAN implementation in Xu et al. [21]; Vanilla Diffusion equals RareGraph-Synth with $\lambda = 0$.

### D. Evaluation Metrics

*1) Fidelity:* Distributional realism was quantified with maximum mean discrepancy (MMD) using a Gaussian kernel and unbiased U-statistic estimator [11]. We report MMD separately for (i) categorical token counts and (ii) continuous lab-value distributions. In addition, a GRU classifier trained on synthetic data and tested on real data assessed downstream utility; balanced-accuracy degradation $< 5\%$ indicates high fidelity.

*2) Privacy:* We adopted a black-box membership inference attack (MIA) in line with Shokri et al. [12] and the density-based DOMIAS attacker proposed by van Breugel et al. [13]. Attackers receive only synthetic records and a small shadow set (5 % of real patients) to calibrate decision thresholds, reflecting a realistic release scenario. Performance is reported as ROC AUC (AUC = 0.5 = random guess).

### E. Temporal Validation with Wearable Signals

To demonstrate temporal generalizability beyond hospital EHR, we fine-tuned RareGraph-Synth on the open SeizeIT2 wearable EEG/ECG dataset (886 seizures, 125 subjects). Phenotype–time edges from seizure annotations were injected into the KG, and the same evaluation pipeline was rerun. Results appear in §6.

### F. Computational Budget

Training RareGraph-Synth for 150 K diffusion steps required 14.6 h (A100, mixed precision). Baseline RCGAN needed 9.8 h; Vanilla Diffusion needed 13.9 h. Memory

TABLE II. COMPARATIVE PERFORMANCE ON THE ALS ULTRA-RARE COHORT (MEAN ± S.D. OVER 5 SEEDS).

| Model | Cat-MMD ↓ | Cont-MMD ↓ | ΔBal-Acc [a] (%) ↓ | MIA AUC ↓ |
|---|---|---|---|---|
| RareGraph-Synth | 0.031 ± 0.004 | 0.028 ± 0.005 | 4.7 | 0.53 |
| Vanilla Diffusion | 0.052 ± 0.006 | 0.051 ± 0.008 | 9.9 | 0.61[b] |
| RCGAN | 0.084 ± 0.012 | 0.079 ± 0.011 | 14.8 | 0.71[b] |
| medGAN | 0.095 ± 0.015 | 0.092 ± 0.014 | 15.7 | 0.74[b] |

[a]. Absolute drop in balanced accuracy of a GRU trained on synthetic data and tested on real.

[b]. Indicates MIA AUC significantly higher than RareGraph-Synth (paired t-test, p < 0.05)

footprint stayed below 18 GB, enabling replication on consumer GPUs.

## VI. RESULTS

Our quantitative and qualitative analyses corroborate that RareGraph-Synth delivers the best fidelity–privacy trade-off among all evaluated models. Key numbers are reported in Table 2 (EHR cohort).

### A. Fidelity to Real EHR Distributions

RareGraph-Synth achieves the lowest Maximum Mean Discrepancy (MMD) across both categorical and continuous code spaces—40 % lower than unguided diffusion and >60 % lower than GAN baselines ($p < 0.01$, Wilcoxon signed-rank). MMD is a kernel two-sample statistic widely used to quantify distributional similarity [11]. Down-stream GRU classifiers trained on RareGraph-synthetic records retain 95.3 % of real-data balanced accuracy, whereas medGAN drops below 85 %. These improvements stem from KG meta-path conditioning, which injects domain priors absent in vanilla diffusion [8] or earlier GANs [5].

*B. Privacy Risk via Membership-Inference*

We measured re-identification vulnerability with two black-box membership-inference attacks (MIAs): the shadow-model attack of Shokri et al. [12] and the density-based DOMIAS attack [13]. RareGraph-Synth yields an average AUROC of 0.53 (Table 2), only 0.03 above random guess (0.50). In contrast, GAN baselines leak substantially more information (AUC ≥ 0.71), echoing prior observations of privacy leakage in medGAN [18]. KG ablation ($\lambda = 0$) raises MIA AUC from $0.53 \pm 0.01$ to $0.61 \pm 0.02$ (paired t, $p<0.05$, *n*=5 seeds), confirming that knowledge guidance reduces over-fitting by smoothing low-density regions.

*C. Ablation and Sensitivity*

*1) Noise-Scale Ablation:* Reducing KG influence ($\lambda$ in Eq. 2) monotonically worsens Cat-MMD (Spearman $\rho = 0.92$) and increases MIA AUC, underscoring the utility of meta-path modulation.

*2) Score-Network Depth:* Increasing Transformer blocks from 3 to 6 yields <2 % Cat-MMD gain but costs 38 % more compute; we therefore retain three blocks for fair comparison.

*3) $\lambda$ sweep:* Setting $\lambda = 0.5, 0.3, 0.1, 0$ produces Cat-MMD {0.028, 0.031, 0.037, 0.052} and MIA AUC {0.52, 0.53, 0.57, 0.61}. The trend confirms that stronger KG guidance improves fidelity while lowering privacy risk; $\lambda = 0.3$ yields the best privacy-utility balance.

*D. Temporal Generalization (SeizeIT2)*

When fine-tuned on the wearable seizure dataset, RareGraph-Synth maintains F1 $\approx$ 0.81 for seizure-event prediction when a CNN is trained purely on synthetic signals and evaluated on held-out real recordings. GAN baselines fall below 0.63. These results show that KG-conditioned diffusion extrapolates well to high-frequency time series and preserves the stability generally reported for diffusion approaches.

Across all fidelity and privacy metrics, RareGraph-Synth either matches or surpasses the evaluated baselines (medGAN, RCGAN, Vanilla Diffusion) while addressing privacy gaps left untouched by earlier diffusion work. The findings validate the central hypothesis that knowledge-guided noise scheduling (Fig. 2) simultaneously boosts utility and confidentiality.

## VII. DISCUSSION & LIMITATIONS

Our results demonstrate that knowledge-guided diffusion reconciles two long-standing goals in ultra-rare-disease analytics—high data fidelity and strong privacy protection. By modulating the noise schedule with meta-path scores from Orphanet, PrimeKG, and other KGs, RareGraph-Synth captures higher-order correlations (e.g., disease–drug–AE motifs) that unconditioned models miss, while reducing membership-inference attack (MIA) success to near-random levels (§6). This aligns with independent diffusion studies that report utility gains from domain priors. Although we validated on two Orphanet conditions and one wearable cohort, ultra-rare disorders can display idiosyncratic care pathways not fully covered by the public Orphadata catalogue [1]. Extending evaluation to disorders with prevalence <1:100000 and incorporating additional textual evidence (e.g., case reports) remain open challenges.

## VIII. CONCLUSION

RareGraph-Synth is the first diffusion model that embeds rare-disease knowledge-graph meta-paths directly into the noise schedule, yielding synthetic EHR trajectories with 40 % lower Cat-MMD than unguided diffusion while remaining near-random to DOMIAS attacks (AUC $\approx$ 0.53), thereby enabling safer data sharing for ultra-rare-disease research. Future work will pair this knowledge-guided schedule with formal differential privacy and broaden evaluation to additional ultra-rare cohorts.

## REFERENCES

[1] S. Weinreich *et al.*, "Orphanet: a European database for rare diseases," *Nucleic Acids Research*, vol. 36, pp. D513–D518, 2008.

[2] S. Köhler *et al.*, "The Human Phenotype Ontology in 2017," *Nucleic Acids Research*, vol. 45, pp. D865–D876, 2017.

[3] Q. Zhu *et al.*, "An integrative knowledge graph for rare diseases (GARD KG)," *Journal of Biomedical Semantics*, vol. 11, no. 1, p. 13, 2020.

[4] Z. Wang *et al.*, "PrimeKG: a multimodal knowledge graph for precision medicine," *Scientific Data*, vol. 10, p. 38, 2023.

[5] E. Choi, S. Biswal, B. Malin, W. F. Stewart, and J. Sun, "Generating multi-label discrete patient records using GANs," in *Proc. Machine Learning for Healthcare* (MLHC), 2017.

[6] M. K. Baowaly, A. Lin, H. Liu, and Y. Zhang, "Synthesizing electronic health records using improved GANs," *Journal of the American Medical Informatics Association*, vol. 26, no. 3, pp. 228–241, 2019

[7] D. P. Kingma and M. Welling, "Auto-Encoding Variational Bayes," arXiv:1312.6114, 2014.

[8] J. Ho, A. Jain, and P. Abbeel, "Denoising diffusion probabilistic models," in *Advances in Neural Information Processing Systems 33* (NeurIPS), pp. 6840–6851, 2020.

[9] Y. Song, J. Sohl-Dickstein, D. P. Kingma, A. Kumar, S. Ermon, and B. Poole, "Score-based generative modeling through stochastic differential equations," in *Proc. International Conference on Learning Representations* (ICLR), 2021.

[10] U.S. Food and Drug Administration, "FDA Adverse Event Reporting System (FAERS) Public Dashboard," 2024. [Online]. Available: https://fis.fda.gov/extensions/FPD-QDE-FAERS/FPD-QDE-FAERS.html

[11] A. Gretton, K. M. Borgwardt, M. J. Rasch, B. Schölkopf, and A. Smola, "A kernel two-sample test," *Journal of Machine Learning Research*, vol. 13, pp. 723–773, 2012.

[12] R. Shokri, M. Stronati, C. Song, and V. Shmatikov, "Membership inference attacks against machine-learning models," in *Proc. IEEE Symposium on Security and Privacy*, 2017.

[13] B. van Breugel, H. Sun, Z. Qian, and M. van der Schaar, "Membership inference on synthetic data," in *Proc. Artificial Intelligence and Statistics* (AISTATS), vol. 206, pp. 3493–3514, 2023.

[14] M. Arjovsky, S. Chintala, and L. Bottou, "Wasserstein GAN," arXiv:1701.07875, 2017.

[15] Y. Ohtsubo, T. Matsukawa, E. Suzuki, "Harnessing GAN with Metric Learning for One-Shot Generation on a Fine-Grained Category," ICTAI 2019, pp. 915-922.

[16] C. Dwork, "Differential privacy," in *Proc. International Colloquium on Automata, Languages and Programming* (ICALP), pp. 1–12, 2006.

[17] I. Goodfellow *et al.*, "Generative adversarial nets," in *Advances in Neural Information Processing Systems 27*, pp. 2672–2680, 2014.

[18] J. Hayes, L. Melis, G. Danezis, and E. De Cristofaro, "LOGAN: Membership inference attacks against generative models," in *Proc. ACM Conference on Computer and Communications Security* (CCS), 2019.

[19] A. Salem *et al.*, "ML-Leaks: Model and data agnostic membership-inference attack," in *Proc. Network and Distributed System Security Symposium* (NDSS), 2019.

[20] A. M. Alaa, B. van Breugel, E. Saveliev, and M. van der Schaar, "How faithful is your synthetic data?," in *ICML Workshop on the Role of ML in Transforming Synthetic Datasets*, 2022.

[21] L. Xu, M. Skoularidou, A. Cuesta-Infante, and K. Veeramachaneni, "Modeling tabular data using conditional GAN," in *Proc. IEEE International Conference on Data Mining* (ICDM), 2019.

[22] N. Swaminathan *et al.*, "Synthetic patient vignettes for clinical simulation," *Journal of Medical Internet Research*, vol. 22, no. 5, e18023, 2020.

[23] R. Torkzadehmahani, P. Kairouz, and B. Paten, "DP-CGAN: Differentially private conditional GAN," in *Proc. CVPR Workshops*, 2020.

[24] T. Qin, X. Gao, J. Zhao, K. Ye, "Destruction-Restoration Suppresses Data Protection Perturbations against Diffusion Models," ICTAI 2023, pp. 586-594.

[25] R. Gatta et al., "A Framework for Event Log Generation and Knowledge Representation for Process Mining in Healthcare," ICTAI 2018, pp. 647-654.

[26] S. Toro and N. Vasilevsky, "The MONDO Disease Ontology: unifying diseases for the world," *medRxiv*, 2022.

[27] A. Rath *et al.*, "The Orphanet Rare Disease Ontology (ORDO)," Orphanet Technical Report, 2015.

[28] G. Stoppa *et al.*, "SeizeIT2: A multimodal dataset of wearable EEG and ECG for epilepsy," *Data in Brief*, vol. 48, 109020, 2024.

[29] National Library of Medicine, "Unified Medical Language System (UMLS)," 2023.